\documentclass[a4paper,twoside]{article}

\usepackage{epsfig}
\usepackage{subcaption}
\usepackage{calc}
\usepackage{amssymb}
\usepackage{amstext}
\usepackage{amsmath}
\usepackage{amsthm}
\usepackage{multicol}
\usepackage{pslatex}
\usepackage{apalike}
\usepackage{algorithm2e}
\usepackage[bottom]{footmisc}
\usepackage{SCITEPRESS}     % Please add other packages that you may need BEFORE the SCITEPRESS.sty package.

\begin{document}

\title{Analysis of 3D Urticaceae Pollen Classification using Deep Learning Models}

\author{\authorname{Tijs Konijn*\sup{1}, Imaan Bijl*\sup{1}, Lu Cao\dag\sup{1} and Fons Verbeek\dag\sup{1}}
\affiliation{\sup{1}Leiden Institute of Advanced Computer Science, Leiden University, Einsteinweg 55, Leiden, The Netherlands}
\email{l.cao@liacs.leidenuniv.nl, \{t.j.e.konijn, e.t.bijl\}@umail.leidenuniv.nl, f.j.verbeek@liacs.leidenuniv.nl}
\textit{*These authors contributed equally to this study.}
\textit{\dag These authors are shared corresponding authors.}
}


\abstract{Due to the climate change, hay fever becomes a pressing healthcare problem with an increasing number of affected population, prolonged period of affect and severer symptoms. A precise pollen classification could help monitor the trend of allergic pollen in the air throughout the year and guide preventive strategies launched by municipalities. Most of the pollen classification works use 2D microscopy image or 2D projection derived from 3D image datasets. In this paper, we aim at using whole stack of 3D images for the classification and evaluating the classification performance with different deep learning models. The 3D image dataset used in this paper is from Urticaceae family, particularly the genera Urtica and Parietaria, which are morphologically similar yet differ significantly in allergenic potential. The pre-trained ResNet3D model, using optimal layer selection and extended epochs, achieved the best performance with an F1-score of 98.3\%}

\onecolumn \maketitle \normalsize \setcounter{footnote}{0} \vfill

\section{\uppercase{Introduction}}
Climate change results in warmer temperature which induces higher amount of pollen production and longer period of pollen season \cite{DAmato2020}. Pollen of different species vary in their allergenic properties. The measured amount of pollen further triggers severer allergic effect and becomes increasingly concerned in healthcare worldwide. Since pollen grains of various species possess different allergenic properties, monitoring the species and the amount of airborne pollen brings in-depth insight in its correlation with the severity of symptoms that hay fever patients experience \cite{Li2023}. Consequently, pollen classification is a crucial step for airborne pollen monitoring. In palynology, pollen classification is mainly performed by highly skilled specialists. They manually classify and quantify pollen grains using light microscopy \cite{Barnes2023}. It is a time-consuming and sometimes error-prone process since the pollen grains of different species may show only subtle morphological differences \cite{Li2023}. An automated pollen classification method is very much needed to increase the efficiency and accuracy. \\
In recent years, machine learning and deep learning techniques have been used to tackle the challenge of pollen classification. The research focus has been on 2D classification. However, pollen grain is 3D object by its nature. 3D classification models are scarcely used in this field. When the specialists manually identify pollen, they also go through the field of depth in order to obtain a 3D impression of the pollen. In this paper, we explore the possibility of using 3D pollen images represented by a stack of 2D images using widefield microscopy for pollen classification and evaluate the classification performance between different deep learning models.\\

\section{\uppercase{Related Work}}
\label{related work}
There is a number of studies using state-of-the-art deep learning classification models for pollen classification. Rostami et al. \cite{rostami-2023} applied various pre-trained CNN models, including AlexNet \cite{10.1145/3065386}, VGG16 \cite{https://doi.org/10.48550/arxiv.1409.1556}, MobileNetV2 \cite{https://doi.org/10.48550/arxiv.1801.04381}, ResNet \cite{https://doi.org/10.48550/arxiv.1512.03385}, ResNeSt \cite{https://doi.org/10.48550/arxiv.2004.08955}, SE-ResNeXt \cite{hu2019squeezeandexcitation}, and Vision Transformer (ViT) \cite{https://doi.org/10.48550/arxiv.2004.089553}, to classify pollen grains from the Great Basin Desert, Nevada, USA. The dataset consisted of 10,000 images of 40 pollen species, and the ResNeSt-110 model achieved the highest accuracy of 97.24\%. They obtained Z-stack images to capture spacial details of pollen grains at different focus level. But the input is still a 224x224 sized 2D image. Daood et al. used a seven layer CNN model to classify 30 pollen types with a rather small dataset: 1,000 samples from light microscopy and 1,161 from scanning electron microscopy (SEM). They achieved a classification rate of approximately 94\%\cite{daood-2016}. Astolfi et al. provided a public annotated pollen image dataset with 73 pollen categories from the Brazilian Savanna \cite{astolfi-2020}. Eight different CNN models are compared to setup a baseline for pollen classification. These CNN models include DenseNet-201 \cite{https://doi.org/10.48550/arxiv.1608.06993}, Inception-ResNet-V2 \cite{https://doi.org/10.48550/arxiv.1602.07261}, Xception \cite{https://doi.org/10.48550/arxiv.1610.02357}, Inception-V3 \cite{https://doi.org/10.48550/arxiv.1512.00567}, VGG16, VGG19, ResNet50 and NasNet \cite{https://doi.org/10.48550/arxiv.1707.07012} . The results showed that ResNet50 with an accuracy of 94.0\% and DenseNet-201 with an accuracy of 95.7\% performed best. Zu et.al proposed a SwinTransformer \cite{https://doi.org/10.48550/arxiv.2103.14030} based model for pollen classification \cite{Zu2024}. In order to diminish the introduced blurry effect during image resizing, Enhanced Super-Resolution Transformer (ESRT) was used to improve the sharpness of the resized images before feeding them into SwinTransformer. The training data consisted of 2D microscopic pollen images stained with Fuchsin. During the experiments, they compared their model with ViT, F2T-ViT \cite{Duan2022}, Efficient-NetV2 \cite{https://doi.org/10.48550/arxiv.2104.00298}, ConvNeXt \cite{https://doi.org/10.48550/arxiv.2201.03545}, ResNet50, and ResNet34. When using their model on pre-trained weights, they achieved highest F1-scores and accuracies for both local and public datasets. \\
Most of the previously mentioned works, however, focus on 2D pollen classification. Several works have presented to make a good use of spacial information captured by z-stack microscopy images. Gallardo et al. designed a multifocus pollen detection model to both localize and classify pollen grains in one go \cite{gallardo-2024}. They used object detection models from Detectron library \cite{wu2019detectron2} to identify pollen for each layer in a z-stack and used a decision algorithm to assign a pollen grain to a class. The two deep learning models they have tested are RetinaNet \cite{https://doi.org/10.48550/arxiv.1708.02002} and Faster R-CNN \cite{https://doi.org/10.48550/arxiv.1506.01497}. The system was trained to recognize 11 pollen types acquired from light microscopy. They were able to locate the pollen with 97.6\% success and these pollen could be classified with an accuracy of 96.3\%. Polling et al. used deep learning model to classify pollen at genus level in the Urticaceae family where it hardly can be distinguished \cite{Polling2021}. They captured Z-stacks with 20 layers of the pollen grains and simplified the 3D information into a 2D image through projection techniques to extract most important feature from z-stack images. These projections included Standard Deviation, Minimum Intensity and Extended Focus. Several classification models, such as VGG16, MobileNetV1 and MobileNetV2, were compared in the experiment. The best performing models were MobileNetV2 and VGG19 reached accuracies of 98.30 and 98.45 respectively using 10-fold cross-validation. Li et al. further compared machine learning and deep learning-based methods for classifying pollen from the Urticaceae family \cite{Li2023}. Machine learning methods required the selection of hand-crafted features. These features were selected based on the observable biological differences between the species, including size, shape and texture. The three different projections were fed into the models as separate channels. The deep learning methods generally outperformed the machine learning methods. The highest-scoring deep-learning method was ResNet50 with 0.994 accuracy on 10-fold cross-validation. This model was closely followed by VGG19 with 0.986 accuracy and MobileNetV2 with 0.985 accuracy.\\

In this paper, we used the raw cropped z-stack as the input of the model, without applying decision algorithm or projection. Our hypothesis is that, by using the z-stack images as a whole, a minimal amount of information is lost. Additionally, the sequence order of z-stack images may contain information that would be lost during projecting the z-stack into 2D space. 

\section{\uppercase{Method}}

\subsection{Data collection}
\begin{figure}[!htbp]
\begin{center}
\includegraphics[height=5cm]{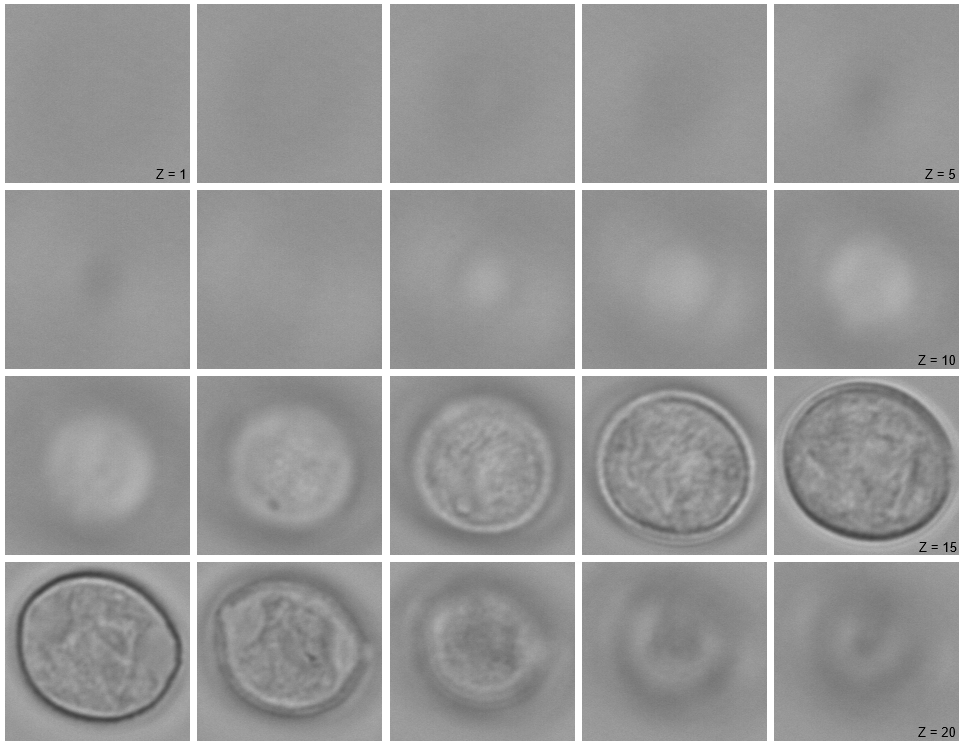}
\end{center}
\caption{A z-stack example of pollen grain. The z-stack contains 20 z-layers, with z-layer 1 on top left and z-layer 20 on bottom right.}
\label{fig: 3D pollen stack}
\end{figure}

The dataset is originally derived from the work of Polling et al \cite{Polling2021,Li2023}. The data is kept in its 3D form that is 20 images in a z-stack taken from different focus levels, as shown in figure \ref{fig: 3D pollen stack}. The dataset incorporates five species from two main genera of Urticaceae family. These are Urtica and Parietaria. The two Species (Urtica urens, Urtica dioica) from Urtica are grouped as the first class and they are all low in allergic profile. However, The two species (Parietaria judaica, Parietaria officinalis) from Parietaria, as the second class, are highly allergic and are the main cause of hay fever and asthma. These two genera are hard to distinguish from microscopy and, therefore, challenging for image pollen classification. The dataset also includes a third class that is a specie from the genus Urtica called Urtica membranacea. The reason for treating it as a single class is because Urtica membranacea is morphologically highly distinguishable from other species belonging to aforementioned two genera. In total, 6472 pollen grains were collected and scanned. Each class has about 2000 pollen grains from collection of wild fresh plants so as to balance classes.
\subsection{Preprocessing}
\begin{figure}[!htbp]
\begin{center}
\includegraphics[height=3cm]{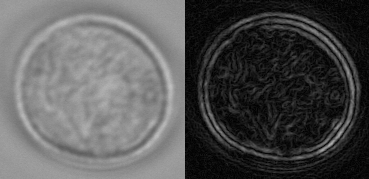}
\end{center}
\caption{Example of edge detection on pollen grain. Figure on the left side is the normal one layer image of a pollen grain. On the right side is the result of edge detection.}
\label{fig: edge detection}
\end{figure}

The z-stack images have the possibility to gain a better classification accuracy because of the spacial dependencies that can be captured in z direction. However, it also causes problems since some layers in z-stack are out of focus. They may not contain much useful information. In addition, the pollen grains reside at different depth level with varied size in z direction. It is not easy to define the focused subset of the stack. Therefore, we firstly defined the central layer of the stack as the layer that has the best focus level. In our case, the sharpest image is found by measuring edge strength that is derived from gradient magnitude of Canny edge detection \cite{Xuan2017} as shown in Figure \ref{fig: edge detection}.

Edge detection makes the central layer easy to find. Different subset layer numbers are tested in this paper to evaluate the change of classification performance with different layer numbers.\\
The z-stack images in the dataset do not all have the same size. We firstly sorted the images and obtained the biggest image size. We used the biggest image size of 224x224 pixels as the input size for deep learning models. In order to import smaller images into deep learning models with the same size, padding step is applied. Average grayscale value of the image is used to pad the edges of smaller images as shown in Figure \ref{fig: edge detection}. The reason of choosing mean padding instead of zero padding is to eliminate the influence of introduced edge artifacts to the performance of pollen classification. 

\subsection{Training}
Since the pollen grains were collected from known species, each palynological reference slide contains pollen grains from a single specie. Thus, the ground truth labels were created right after the microscopy scanning. For training, the data is firstly split into training, validation, and test sets. 10 \% of the data is kept as test set. We used 10-fold cross-validation to evaluate the classification performance. At every fold, a new training and validation split is used. The training data is processed further with data augmentation while the validation set is not processed. The train/validation ratio is 9 to 1. The data augmentation is performed by vertically and horizontally flipping the images to increase the diversity of the training dataset. Only horizontal and vertical flips were performed so as to minimize the artifact introduction to the training dataset.\\
We tested five 3D deep learning models for this project. They are ResNet3D \cite{https://doi.org/10.48550/arxiv.1812.03982}, MobileNetV2, SwinTransformer 3D \cite{yang2023swin3d}, the classification branches of RetinaNet \cite{https://doi.org/10.48550/arxiv.1708.02002} and Faster R-CNN \cite{https://doi.org/10.48550/arxiv.1506.01497}. 18 layer Resnet3D model shows promising pollen classification results in previous studies cf. \ref{related work}. The default pre-trained weights were used. The weights were obtained by pre-training the model on two different datasets: the Kinetics \cite{carreira-2017} and Sports-1M \cite{Karpathy_2014_CVPR}. 
MobileNet is known for its light-weighted structure. It has been carefully evaluated for pollen classification \cite{Polling2021,Li2023}. We used MobileNetV2 because it can use different input layer size. We used 3D MobileNetV2 implemented by Köpüklü, et al. \cite{https://doi.org/10.48550/arxiv.1904.02422}.   Swintransformer is a representative model from vision transformers, characterized by its hierarchical structure using shifted windows. In this study, SwinTransformer 3D is used. Two object detection models are included in the evaluation as well since they have been reported to work well for pollen identification and classification. We only tested their classification branches in this study. RetinaNet consists of a feature pyramid network (FPN) backbone on top of a feed forward ResNet architecture. The ResNet architectures used here are ResNet50 and ResNet3D, since they allow for 3D input. Compared to RetinaNet, Faster R-CNN, depending on region proposal network, has a higher identification performance of pollen grains but a bit higher inference time \cite{gallardo-2024}. The same backbones as RetinaNet are used for Faster R-CNN.\\
Pytorch library is used to evaluate 3D deep learning models. Pytorch has ready-made 3D CNN models that can be imported and has data augmentation. Pytorch also allows for GPU acceleration which is necessary for deep learning models that are computationally very expensive. The time that it takes for a model to complete one epoch was measured on a NVIDIA RTX 4070Ti GPU. This GPU has 12GB of VRAM.\\
For all models we use metrics: accuracy, loss, and F1 score of the model to evaluate the performance.

\section{\uppercase{Results}}
\subsection{Optimal Number of layers}
The number of layers in a z-stack were evaluated to observe if it makes a difference in classification performance. We have tested the subsection of layers for 4, 6, 8 and 20 which is the total amount of the z-stack. The results, summarized in table \ref{Tab: ResNet3D different opframes}, showed that the increase of layer numbers do not assure a better classification performance. Increasing the amount of layers only resulted in a 0.02 increase of accuracy from 6 layers to 10 layers. The sudden decline in accuracy at 20 layers and the fact that the model does not stagnate has been a reason to test the model with more epochs. Remainder of the research was performed with 6 optimal layers instead of the 10 layers, because we have shown that improvement is marginal and extra time is needed for training with 4 extra layers. 
\begin{table}[!htbp]
    \begin{center}
    \begin{tabular}{l|l|l|l|l}
    \textbf{Layers} & loss & F1-score & accuracy & Time \\
    \hline
    4 layers& 0.055 & 0.980 & 0.980 & 90 \\
    6 layers & 0.052 & 0.981 & 0.981 & 92 \\
    8 layers & 0.050 & 0.980 & 0.980 & 92 \\
    10 layers & 0.046 & 0.983 & 0.983 & 93 \\
    20 layers & 0.069 & 0.977 & 0.977 & 97
    \end{tabular}
    \end{center}
    \caption{ResNet3D model testing performance with varied number of layers. Time: time for one epoch in seconds.}
    \label{Tab: ResNet3D different opframes}
\end{table}

\subsection{Increasing Amount of Epochs}
% increasing epochs does not have an effect on accuracy
In order to examine if the learning curves from ResNet3D stagnate, 50 epochs of the model with 6 optimal layers was performed. 

\begin{figure}[!htbp]
\begin{center}
\includegraphics[trim={0 0 7.5cm 0}, clip, height=5.5cm]{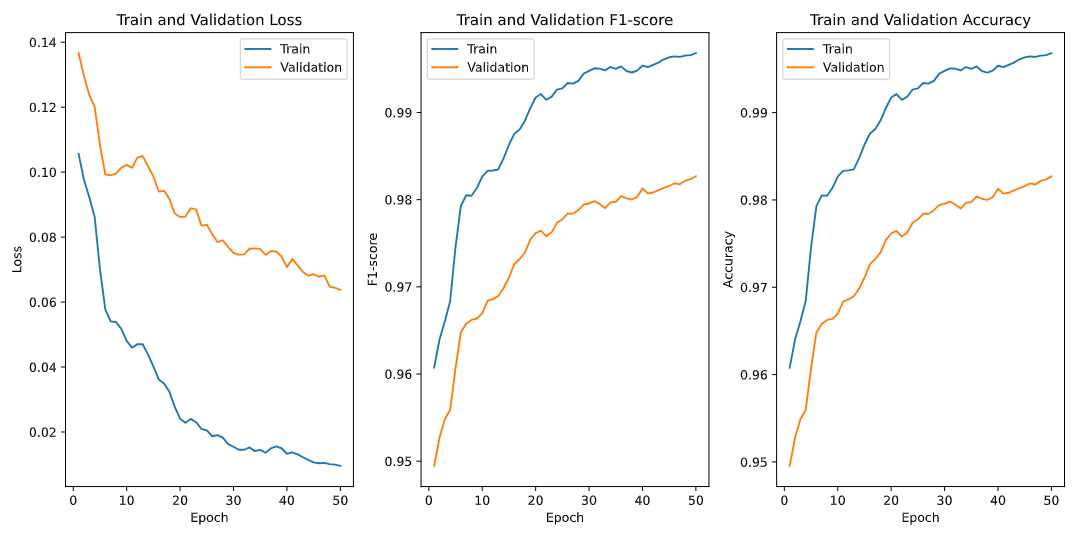}
\end{center}
\caption{These graphs represent the pre-trained ResNet3D model with 6 optimal layers and 50 epochs. On the test set this model achieved a F1-score of 0.982 and a loss of 0.053.}
\label{fig: ResNet3D 50 epochs}
\end{figure}

We can observe from Figure \ref{fig: ResNet3D 50 epochs} that an increase in epochs only marginally improves the accuracy with 0.01 for 20 extra epochs. This can also be seen in table \ref{Tab: ResNet3D 30 vs 50 epochs}. The small increase in accuracy makes the trade-off with 20 extra epochs not necessary.

\begin{table}[!htbp]
    \begin{center}
    \begin{tabular}{l|l|l|l}
    \textbf{ResNet3D} & loss & F1-score & accuracy \\
    \hline
    30 epochs & 0.052 & 0.981 & 0.981  \\
    50 epochs & 0.053 & 0.982 & 0.982  \\
    \end{tabular}
    \end{center}
    \caption{ResNet3D model testing performance difference between 30 and 50 epochs}
    \label{Tab: ResNet3D 30 vs 50 epochs}
\end{table}

\subsection{Comparison of Deep Learning Models}
During the comparison, we have given attention to the performance as well as the efficiency of the models. We need a model that is efficient and precise to classify pollen from large amount of microscopy images. For each model, we start from the hyperparamters that are the optimal set for ResNet3D as shown in Table \ref{hyperparameters}. We adjust the hyperparamters accordingly in order to find the optimal hyperparamters for each model in our comparison.
\begin{table}
    \centering
    \begin{tabular}{c|c}
    hyper-parameter & value \\ \hline
       folds  & 10 \\ 
       epochs  & 30 \\
    learning rate & 0.0001 \\
    augmentation threshold & 0.5 \\
    number of layers & 6 \\ 
    training batch size & 16 \\ 
    validation batch size & 16 \\ 
    \end{tabular}
    \caption{hyperparameters setting}
    \label{hyperparameters}
\end{table}

The ResNet3D was evaluated on pre-trained and non pre-trained situation. Compared to the non pre-trained model, the pre-trained model performs significantly better with a F1-score of 98.1\%. The time for both models to train was almost the same, around 92 seconds per epoch. For MobileNetV2 3D, we adjusted the hyper-parameters for learning rate ([0.001, 0.0001, 0.00001]), augmentation threshold ([0.1, 0.2, 0.5, 0.7]) and batch size ([16, 20, 50, 70]). We observed that the hyper-parameters listed in Table \ref{hyperparameters} are the optimal parameters for MobileNetV2 3D training from scratch. However, the hyperparameters slightly changed for pre-trained MobileNetV2 3D, where learning rate is 0.001, augmentation threshold is 0.2 and batch size for validatation is 20.\\
For the 3D Swintransformer model, the hyper-parameters for the number of attention heads and window size were used default. There were no pre-trained weights available. Therefore, the Swintransformer 3D was trained from scratch. We adjusted learning rate, augmentation threshold, batch size in the same ranges to obtain the best performance. At the end, we used the same learning rate and augmentation threshold but different batch size of 50 for training and 20 for validation.\\ 
Faster R-CNN was trained on the ResNet50 backbone first. However, it took 10 minutes to train one epoch. Compared to 100 seconds to train for ResNet3D model in Table \ref{Tab: ResNet3D different opframes}, it took much longer time and resulted in training five hours for one fold out of 10-Fold cross validation. Furthermore, this model was only able to achieve an F1-score of 94.3\% on the test set after 30 epochs. It needs much more epochs before the results achieve the same performance as ResNet3D. Therefore, Faster R-CNN with ResNet50 is not included in the following comparison. Faster R-CNN model with ResNet3D backbone is a lot better compared to the ResNet50 backbone version. The model achieved a F1-score of 0.979, an accuracy of 0.979 with a loss of 0.072. The model took 115 seconds per epoch to train. We used the same hyperparameters as ResNet3D.\\
The last model to be tested was RetinaNet. First a pre-trained ResNet50 backbone model was compared to a non pre-trained model. Both of the models stagnate after 20 epochs, however the pre-trained model has a significantly higher F1-score. Subsequently, it was tested if an increase in the total epochs would make a difference for the ResNet50 backbone. The only difference that we observed is the decrease of loss. Therefore, it is not necessary for the model to be trained with 50 epochs. We also compared between different backbones for RetinaNet. Unlike the Faster R-CNN model, RetinaNet with ResNet50 backbone was a lot faster. Therefore a comparison can be made as shown in Table \ref{Tab: Model comparison}.
\begin{table}[!htbp]
    \begin{center}
    \begin{tabular}{l|l|l|l}
    Model & loss & F1-score & accuracy\\
    \hline
    ResNet3D & 0.112 & 0.959 & 0.959 \\
    ResNet3D* & 0.052 & 0.981 & 0.981\\
    MobileNetV2 &  0.131 & 0.959 & 0.959 \\
    MobileNetV2* & 0.073 & 0.974 & 0.974 \\
    Swintransformer 3D & 0.286 & 0.924 & 0.924 \\
    \hline
    Faster R-CNN with & & & \\
    \hline
    RestNet3D* & 0.072 & 0.979 & 0.979 \\
    \hline
    RetinaNet with & & &  \\
    \hline
    ResNet50 & 0.151 & 0.954 & 0.954  \\
    ResNet50* & 0.134 & 0.965 & 0.965  \\
    ResNet3D* & 0.060 & 0.978 & 0.978  \\

    \end{tabular}
    \end{center}
    \caption{Classification performance comparison between models.  * means Pre-trained model.}
    \label{Tab: Model comparison}
\end{table}

\section{\uppercase{Discussion and Conclusion}}
The primary objective of this study was to enhance the classification accuracy of pollen in Urticaceae family using deep learning techniques, specifically leveraging 3D convolutional neural networks (3D CNNs). This was motivated by the need for precise identification of pollen grains due to their varying allergenic potential, which is crucial for environmental and health monitoring.\\
The implementation of edge detection to identify the best focal point significantly contributed to the improved classification accuracy. By focusing on the image layer with the highest intensity change, the model could utilize the most informative features of the pollen grains. This preprocessing step mitigated the issue of blurry layers and ensured that the input to the neural network was of the good quality, thereby enhancing the model's ability to learn and distinguish between different pollen types.\\
Incremental improvements in model performance were observed with the optimization of layer selection and an increase in the number of epochs. However, the enhancements were marginal, suggesting that the models had already achieved near-optimal performance using the given list of hyperparameter settings as shown in Table \ref{hyperparameters}.  Through empirical evaluation, six layers were chosen as the optimal subset for further analysis. This decision was based on the trade-off between performance gain and computational efficiency. The results showed that increasing the number of layers from six to ten yielded only a marginal improvement in F1-score and accuracy (from 98.1\% to 98.3\%), while 22\% extra time was needed for these 4 extra layers.\\
The research further focused on comparing different 3D deep learning models, including ResNet3D, MobileNetV2 3D, Swin Transformer, Faster R-CNN, and RetinaNet, with particular attention to the performance of the ResNet50 and ResNet3D backbones for the object detection models. The models were trained and tested on a dataset consisting of 6472 pollen stack images, each captured with 20 slices along the Z-axis. We conclude that pre-trained ResNet3D alone performs the best with F1-score of 0.981 and a loss of 0.052. Faster R-CNN and RetinaNet with pre-trained RestNet3D are the second and third positions with F1-score of 0.979 and 0.978 respectively. However, the study utilized pre-trained weights for several models, which were derived from datasets like Kinetics \cite{carreira-2017}and Sports-1M \cite{Karpathy_2014_CVPR}. While these weights helped accelerate training and improve accuracy, they are optimized for video or action recognition tasks rather than pollen images. This mismatch might limit the models' ability to fully capture domain-specific features, leaving room for further improvement with domain-specific pre-training.\\
The MobileNetV2 model performs reasonably well, considering its light weight and few parameters. Using MobileNetV2 for pollen classification could be desirable if reducing computational cost is a special requirement. The pre-trained MobileNetV2 outperforms the model trained from scratch by 1.5 percent. Interestingly, the hyper-parameter settings working the best for the model trained from scratch are not those resulting the best for the pre-trained model. The pre-trained model benefits more from a higher learning rate. It is probably because the model already captured general features from a large dataset. Additionally, the model does not suffer from a destabilized learning process, which is often a result of a high learning rate. The pre-trained model is already close to a optimal performance. Therefore, faster adjustments are beneficial \cite{smith2017super}. The MobileNetV2 does not seem to benefit from longer training, suggesting the model over-fits on the training data when trained for 50 epochs. \\
Swin Transformer 3D did not reach to an optimal performance in the evaluation partially because of the lack of pre-trained weights available for this model. However, when the models were trained from scratch, Swin Transformer performs worse than ResNet3D and MobileNetV2.
Swintranformer benefits from larger data size which cannot be met in our case. Its architecture, with complex attention mechanisms and hierarchical processing units, makes it more sensitive to limited dataset. In addition, Swin Trainsformer did not benefit from longer training, suggesting over-fitting when trained on 50 epochs. \\
The ResNet3D backbone demonstrated superior performance compared to the ResNet50 backbone, achieving an F1-score and accuracy of 97.8\% on the test set with RetinaNet. However ResNet3D classification model performed better compared to the object detection and classification models. Resnet3D achieved an F1-score and accuracy of 98.3\% when trained with 10 optimal layers. This indicates the efficacy of 3D convolutional neural networks (CNNs) in capturing spatial features from stack images, which is critical for distinguishing between morphologically similar pollen grains.\\

The close F1 score and accuracy in our comparison of models indicate a balanced performance across both precision and recall, suggesting that the model is equally effective at correctly identifying true positives and minimizing false positives. Such balance is crucial in applications like pollen classification, where both the presence of pollen and the correct identification of its type are important for accurate environmental monitoring and allergen forecasting.\\
Despite the promising results achieved in this study, the previous research by Chen Li et al. \cite{Li2023} demonstrated slightly better performance. Their approach, which utilized projection technique combined with various deep learning architectures, achieved an accuracy of 99.4\% in classifying Urticaceae pollen. This underscores the potential for further optimization in our methods, possibly by integrating additional preprocessing techniques or refining the neural network architectures to match or surpass the results.

\section{\uppercase{Future Work}}
This research has ample possibilities to work on in the future. Currently the optimal focal layer selection is not always entirely correct. If an optimal focal layer is close to the top or bottom layers and the total amount of subset layers is set too high, it will deviate the focal layer from the central position. This could have an impact on the performance of 3D CNNs. In addition, if we have sufficient computation power in the future, the necessity of finding the optimal focal layers should be evaluated. Therefore, further research should look into the improvement of heuristics and possibility of making better 3D data.\\
In addition, the scope of pollen classification could be expanded to include a wider variety of pollen families beyond Urticaceae. Investigating the application of 3D CNNs to other pollen types could validate the generalizability and robustness of the proposed methodology. This expansion would involve collecting and annotating new datasets from different plant families, which may present unique challenges in terms of morphological diversity and data complexity. Additionally, understanding the specific allergenic properties of various pollen families would further enhance the practical applications of these models in environmental health and allergen forecasting. Instead of widefield microscopy, confocal laser scanning micrsocpy (CLSM) is a good option providing baseline on pollen morphology and nuclei stained with DAPI. 

\bibliographystyle{apalike}
{\small
\bibliography{paper}}

\end{document}